\documentclass[10pt,twocolumn,letterpaper]{article}

\usepackage{wacv}
\usepackage{times}
\usepackage{epsfig}
\usepackage{graphicx}
\usepackage{amsmath}
\usepackage{amssymb}
\usepackage{enumerate}

\wacvfinalcopy 

\ifwacvfinal\fi
\begin{document}

\title{Is Image Super-resolution Helpful for Other Vision Tasks?}
\author{Dengxin Dai\thanks{Joint first author}  \hspace{.4cm} Yujian Wang$^*$ \hspace{.4cm}  Yuhua Chen \hspace{.4cm} Luc Van Gool\\
Computer Vision Lab, ETH Zurich\\
{\tt\small \{dai, yuhua.chen, vangool\}@vision.ee.ethz.ch, yjwang@student.ethz.ch}
}

\maketitle
\ifwacvfinal\thispagestyle{empty}\fi

\begin{abstract}
  Despite the great advances made in the field of image super-resolution (ISR)
  during the last years, the performance has merely been evaluated
  perceptually. Thus, it is still unclear whether ISR is helpful for other
  vision tasks. In this paper, we present the first
  comprehensive study and analysis of the usefulness of ISR for other
  vision applications. In particular, six ISR methods are evaluated
  on four popular vision tasks, namely edge detection, semantic image
  segmentation, digit recognition, and scene recognition. We show that applying ISR to input images of other vision
  systems does improve their performance when the input images are of
  low-resolution. 
We also study the correlation between four standard perceptual evaluation criteria (namely 
PSNR, SSIM, IFC, and NQM) 
and the usefulness of ISR to the vision tasks. 
Experiments show that they correlate well with each other in general,  
but perceptual criteria are still not accurate enough to be used as  full proxies 
for the usefulness. We hope this work will inspire the community to evaluate ISR 
methods also in real vision applications,  and to adopt ISR as a pre-processing step of other
  vision tasks if the resolution of their input images is low.

\end{abstract}


\section{Introduction}
\label{sec:intro}
Image super-resolution (ISR) aims to sharpen smooth rough edges and
enrich missing textures in images that have been enlarged using a
general up-scaling process (such as a bilinear or bicubic process),
thereby delivering an image with high-quality
resolution~\cite{Freeman-CGA-2002, Yang-TIP-2010, Zeyde-CS-2012,
  Timofte-ICCV-2013, Dong-ECCV-2014, JOR:EG15}. ISR systems can be
used to adapt images to displaying devices of different dimensions, to
map image textures to 2D/3D shapes, and to deliver pleasing
visualization for data that are inherently low-resolution such as
image or videos from surveillance cameras.  Despite the popularity of
ISR in the past years, their performance has merely been evaluated
perceptually and/or by evaluation criteria reflecting perceptual
quality such as PSNR and SSIM. Therefore, it is still unclear whether
ISR is helpful in general to other vision tasks and whether the
perceptual criteria are able to reflect the usefulness. 
This paper answers the questions.

We here present reasons why ISR can be helpful for other vision tasks,
in addition to improving perceptual quality. As we know, most of
current vision systems consist in two phases: training and
testing. Although features have been designed to overcome the
influence of scale changes, it is still a blessing if 1) the training
and testing images are of the same/similar resolution; and/or 2)
input images can be converted to the resolution at which the
features and the models were designed. It happens quite common that
training and testing data are of different resolutions, \eg training
images are from expensive sensors while testing images from cheap
ones. If testing images are of higher resolution, down-sampling them
with linear filters does the job. If the opposite holds, however,
sophisticated ISR methods are required to super-resolve the testing
images. Also, vision systems are often designed and optimized (\eg the
features) for images of the most `popular' resolution at the time. ISR
is useful to super-resolve images which are of lower-resolution than
the images for which the features and models are designed and
learned. One example is object recognition with surveillance cameras:
popular features~\cite{Lowe_SIFT, Dalal_HoG, cnnfeatureoffshelf} for object recognition are designed for normal images
which are of higher-resolution than surveillance scenes in
general. For this case, even the training data and testing data
are of the same resolution, ISR is still helpful by
enabling feature extraction at an appropriate resolution.


In order to sufficiently sample the space of ISR methods and potential
vision tasks, six ISR methods are chosen and evaluated on four popular
vision applications. The ISR methods are Zeyde
\etal~\cite{Zeyde-CS-2012}, ANR~\cite{Timofte-ICCV-2013},
A+~\cite{Timofte-ACCV-2014}, SRCNN~\cite{Dong-ECCV-2014}, 
JOR~\cite{JOR:EG15}, and SRF~\cite{SR_forest}.
They are chosen because 1) they are popular and representative; 
2) they have code available; and 3) they are computationally efficient.  
The four vision applications include image boundary detection, semantic image segmentation, 
digit recognition, and scene recognition. The tasks are chosen because they are representatives of
current low- and high-level vision tasks. The data of digits is chosen because low-resolution 
inputs are very likely to occur in this field. For all the tasks, we apply 
standard approaches with varying modes of the input images: from
low-resolution images, to super-resolved images by
the six ISR methods, and to the high-resolution images. The experimental
results suggest that ISR is helpful for these vision
tasks if the resolution of the input images are low, and that the
standard evaluation criteria, including PSNR, SSIM, IFC, and NQM, correlate generally well with
the usefulness of ISR methods, but should not be used as the full proxies of the usefulness 
if high precision is required.

The paper is organized as follows. Section~\ref{sec:relatedwork}
reports related work. Evaluation on the four vision tasks are
conducted in Section~\ref{sec:ed} to Section~\ref{sec:scene}. Finally,
the paper concludes in Section~\ref{sec:conclusion}.

\section{Related Work}
\label{sec:relatedwork}
There is a large body of work addressing image super-resolution
task. We breifly summarize them. 
The oldest direction is represented by variants of interpolation, such
as Bilinear and
Bicubic~\cite{Duchon-JAM-1979,Thevenaz-BOOK-2000}. They represent the
simplest and the most popular methods. However, they often produce
visual artifacts such as blurring, ringing, and blocking, which
follows the fact that their assumptions of smoothness and band-limited
image data hardly hold in real cases. Due to these reasons, more
realistic priors and regularization have been developed, such as the
sparse derivative priors in~\cite{Tappen-WSCTV-2003}, the PDE-based
regularization in~\cite{Tschumperle-PAMI-2005}, the edge
smoothness prior in~\cite{Dai-CVPR-2007}, and gradient profile~\cite{Sun-CVPR-2008}. Despite the improvement by
these methods, the explicit forms of prior are still insufficient to
express the richness of real-world image data.

In recent years, example-based image super-resolution has raised the
most attention due to its good performance and simplicity. In this
stream, the task is to learn a mapping function from a collection of
LR images and their corresponding high-resolution (HR) ones. The LR and HR data can be
collected from the test image itself or from an external dataset.
Methods~\cite{Freedman-TOG-2011, Glasner-ICCV-2009, selfsimaccv, Huang_2015_CVPR}
in the former stream draw on the `self-similarity' of images across scales,
and have obtained great success. However, they are normally relatively slow because on-line learning is needed for the dictionary. 
Methods in the latter group rely on extra training data, unleashing the learning capacity
of many learning methods. The KNN method~\cite{Freeman-CGA-2002}
and its variants~\cite{Chang-CVPR-2004, Yang-ICCV-2013, JOR:EG15, dai:MI}
have gained great attention. More sophisticated learning methods such as Sparse Coding~\cite{Yang-TIP-2010,  Kim-PAMI-2010, Timofte-ICCV-2013, Timofte-ACCV-2014,  deeply:improved},
SVM~\cite{ISR:SVM}, Random Forests~\cite{SR_forest, NBSRF, cSRF}, and Deep Neural Network~\cite{Dong-ECCV-2014, deep:cascade, deeply:improved} have been applied widely to the task as well.
One exceptional work is ~\cite{ISR:internet}, using scene matching with internet images for image super-resolution.
Since example-based methods with extra training data obtain state-of-the-art performance for ISR, our 
evaluation is focused mostly on this stream.


There is also a survey paper on ISR~\cite{ISR:survey}, providing an excellent summary of  
the theory and applications of ISR. \cite{seven:way} exploits 
seven ways to improve the performance of general example-based ISR methods.  
The work most relevant to ours is~\cite{SR:benchmark}, where different
ISR methods are evaluated. While sharing similarities, the two methods
still differ significantly. \cite{SR:benchmark} conducted user
studies for perceptual evaluation, solely with visual comparison and
under evaluation criteria such as PSNR and SSIM. Our work, however, integrates
ISR methods into systems of other vision applications and
evaluates the usefulness of ISR to these vision tasks. 
There are also works employing ISR to improve the quality (resolution) of the input images of other vision algorithms,
such as \cite{face:SR08} for face recognition and \cite{Jing_2015_CVPR}
for pedestrian identification.
However, these tasks are specific and the ISR methods used are highly specialized. Our work, however,
evaluates general ISR methods with a variety of popular vision tasks.

\section{Evaluation}
\label{se:evalucation}

\begin{table*} [tb]{
\centering
\resizebox{\textwidth}{!}
{
\begin{tabular}{|l|c|ccccccc|c|}
  \hline
  \multicolumn{2}{|c|}{BSDS300} & Bicubic & Zeyde~\etal\cite{Zeyde-CS-2012} & ANR\cite{Timofte-ICCV-2013} & SRCNN\cite{Dong-ECCV-2014} & A+\cite{Timofte-ACCV-2014} & JOR\cite{JOR:EG15} & SRF\cite{SR_forest} & Original \\
  \hline
  \hline
  \textbf{$\times$3} & PSNR & 27.15 & 27.87 & 27.88 & 28.10 & \textbf{28.18} & \underline{28.17} & \underline{28.17} & --- \\
  & SSIM & 0.736 & 0.770 & 0.773 & 0.777 & \textbf{0.781} & \textbf{0.781} & \underline{0.780} & --- \\
  & IFC & 2.742 & 3.203 & 3.248 & 3.131 & \textbf{3.374} & 3.360 & \underline{3.366} & --- \\
  & NQM & 27.42 & 31.80 & 31.95 & 31.28 & 32.35 & \textbf{32.41} & \underline{32.40} & --- \\
  \hline
  & AUC & 0.647 & \textbf{0.675} & 0.665 & 0.668 & \textbf{0.675} & \underline{0.674} & \underline{0.674} & 0.696 \\
  \hline
  \textbf{$\times$4} & PSNR & 25.92 & 26.51 & 26.51 & 26.66 & \textbf{26.77} & \underline{26.74} & \underline{26.74} & --- \\
  & SSIM & 0.667 & 0.697 & 0.699 & 0.702 & \textbf{0.709} & \underline{0.707} & \underline{0.707} & --- \\
  & IFC & 1.839 & 2.195 & 2.231 & 2.117 & \textbf{2.325} & \underline{2.316} & 2.293 & --- \\
  & NQM & 21.15 & 24.30 & 24.37 & 24.19 & \textbf{24.98} & \underline{24.96} & \textbf{24.98} & --- \\
  \hline
  & AUC & 0.595 & 0.647 & 0.635 & 0.650 & \textbf{0.656} & \underline{0.655} & 0.652 & 0.696 \\
  \hline
\end{tabular}
}
\caption{Average PSNR, SSIM, IFC, NQM values of ISR methods on BSDS300 and average 
AUC values of boundary detection via CBD~\cite{isola2014crisp} on the super-resolved images by 
the ISR methods and the original images. The best one is shown in \textbf{bold} and the second 
best \underline{underlined}.}
\label{tab:ed}}
\end{table*}

\begin{figure*} [tb]
\centering
\begin{tabular}{cc}
  \includegraphics[trim=15 15 15 15, clip, width=0.5\textwidth]{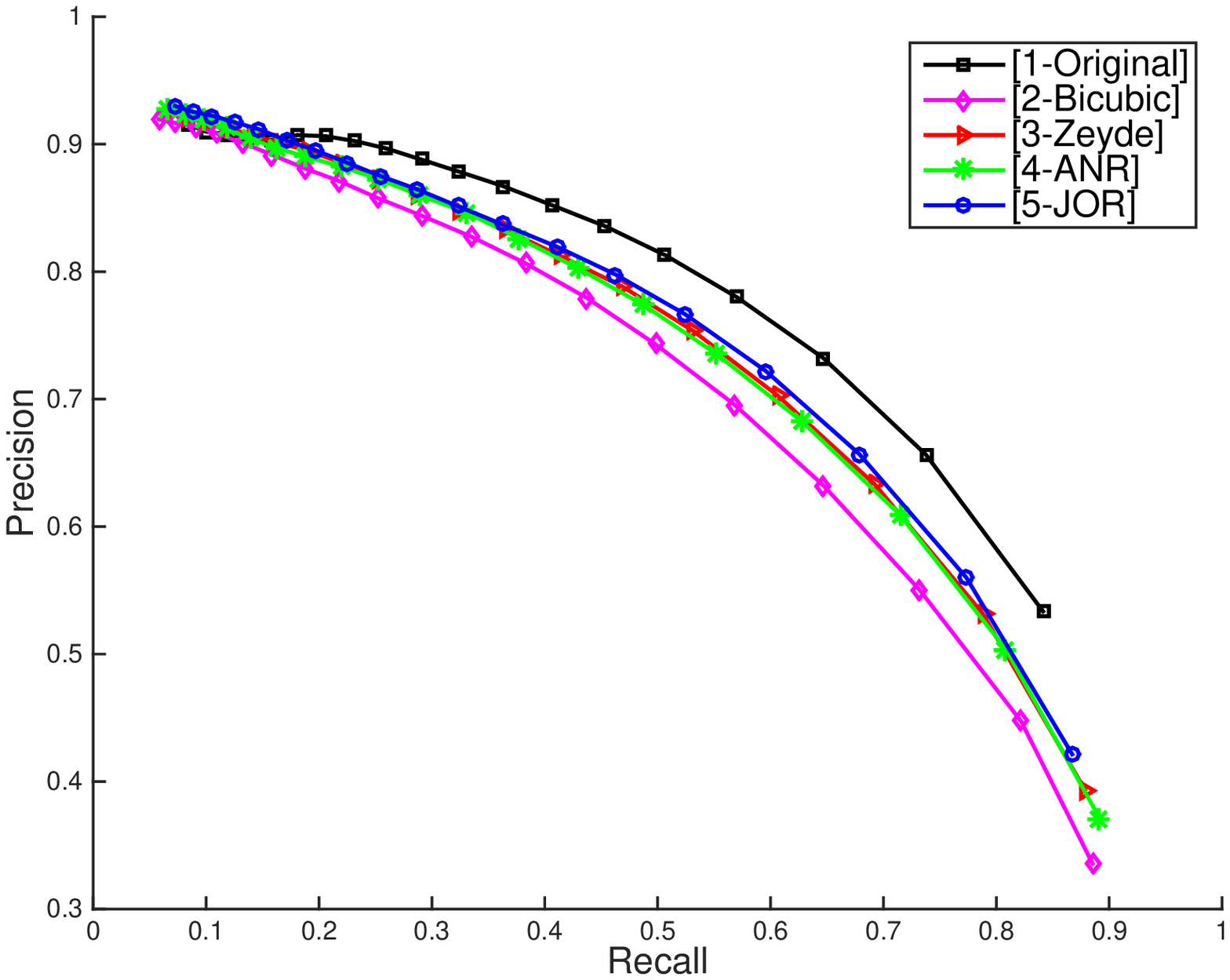} &
  \hspace{-4mm}
\includegraphics[trim=15 15 15 15, clip, width = 0.5\textwidth]{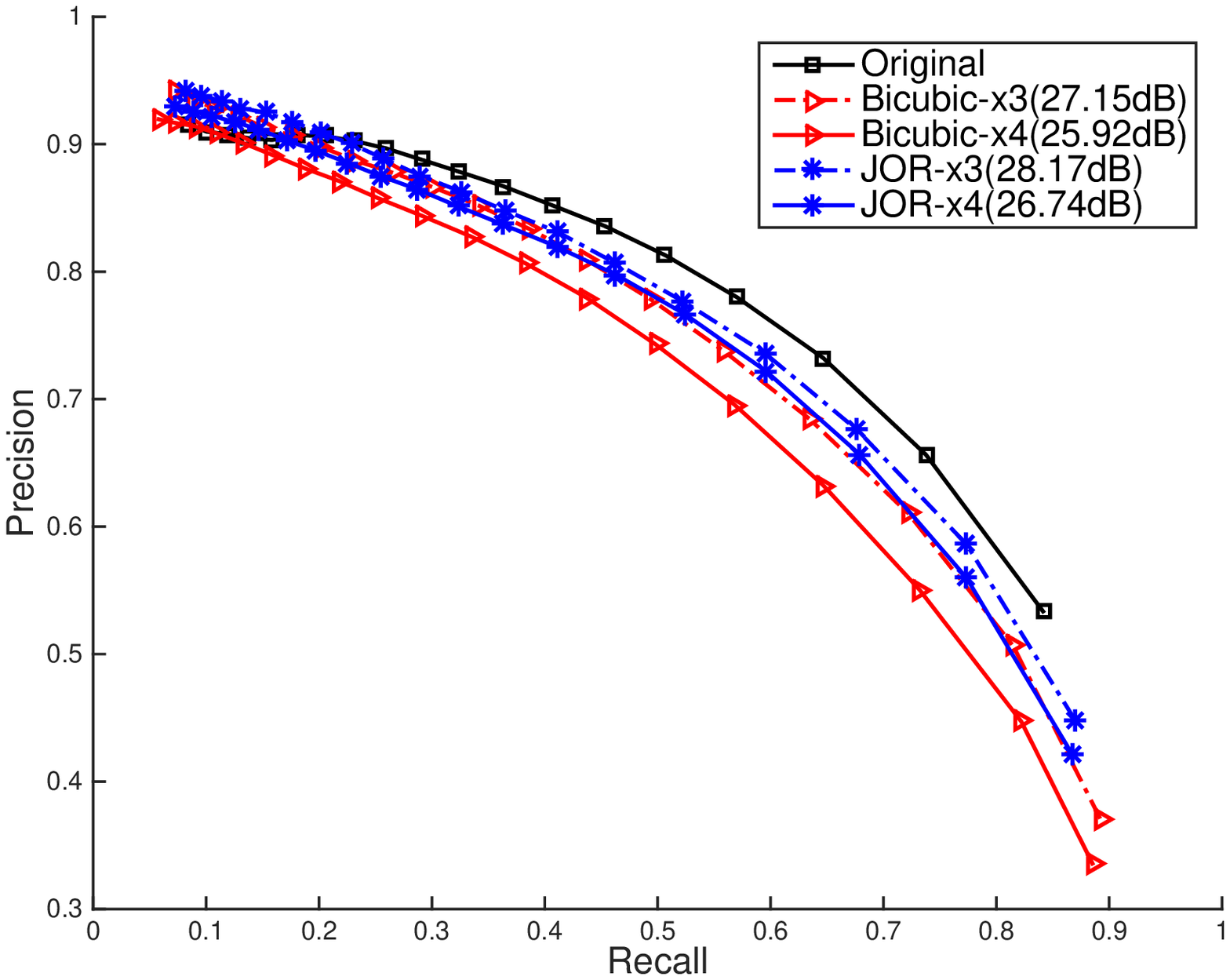} \\
  \footnotesize{\text{(a) PR curves with scaling factor x4}} &
  \footnotesize{\text{(b) PR curves with scaling factor x3 and x4}}
\end{tabular}
\caption{Average PR curves of boundary detection via
  CBD~\cite{isola2014crisp} on the super-resolved images by some of
  the five ISR methods and on the original images of BSDS300.  (a)
  curves for scaling factor x4, where SRCNN and A+ are not shown for
  visual clarity as they are very similar to JOR. (b) a comparison of
  scaling factor x3 and x4, where only Bicubic Interpolation and JOR
  are shown for visual clarity.}
 \label{fig:ed_method}
\end{figure*}

In this section, we briefly describe the six ISR methods: Zeyde
\etal~\cite{Zeyde-CS-2012}, ANR~\cite{Timofte-ICCV-2013},
A+~\cite{Timofte-ACCV-2014}, SRCNN~\cite{Dong-ECCV-2014}, 
JOR~\cite{JOR:EG15}, and SRF~\cite{SR_forest}, followed by the evaluation on the four vision
tasks.  The six methods, starting out with the results of Bicubic
interpolation, learn from examples to recover the missing
high-frequency parts. As to the examples, the six methods are all
trained with the same training dataset from~\cite{Yang-TIP-2010},
which consists of $91$ images of flowers, faces, \etal. For implementation, we use the codes
provided by the authors. Readers are referred to their papers for
details. As to scaling factors, we evaluate with
$\times$3 and $\times$4, which are commonly used in previous papers.

For datasets, we use the standard ones for the four vision tasks, though not the most challenging ones. To generate
inputs for our evaluation, we downscale the original images of the
datasets by factors $\times$3 and $\times4$ to create the
low-resolution (LR) images and then upscale them by each of the six
ISR methods to the resolution of the original images, which are then
used as the inputs for the vision tasks. The standard approaches to
the four tasks are then applied to all the six super-resolved versions of the
images. The corresponding performances are
recorded to evaluate the usefulness of the ISR methods for the vision
tasks, with a comparison to  Bicubic Interpolation, and the
original images. We also evaluate the ISR methods on these datasets
with four standard perceptual criteria~\cite{SR:benchmark}, namely
PSNR, SSIM, IFC, and NQM, in order to see their correlation to the
usefulness of ISR to these vision tasks.


\subsection{Boundary Detection}
\label{sec:ed}

Boundary Detection (BD) is a very popular low-level vision task and
serves as a crucial component for many high-level vision
systems~\cite{Martin-ICCV-2001, isola2014crisp}. This section
evaluates the usefulness of ISR methods for BD.  We use Crisp Boundary
Detection~\cite{isola2014crisp} (CBD), which is an unsupervised
algorithm, deriving an affinity measure with point-wise mutual
information between pixels and utilizing this affinity with spectral
clustering method to detect boundaries. It produces pixel-level
boundaries and achieves state-of-art results.  The performances are
evaluated on the BSDS300 dataset~\cite{Martin-ICCV-2001}. The whole
dataset consists of 300 images (200 for training and 100 for testing)
along with human annotations.  The quality of detected boundaries is
evaluated by precision-recall (PR) curves, following Berkeley
Benchmark~\cite{Martin-ICCV-2001}.

\begin{figure*} [tb]
\setlength{\tabcolsep}{1pt}
\begin{tabular*}{\textwidth}{cccccccc}
    Original & Bicubic & Zeyde~\cite{Zeyde-CS-2012} & ANR\cite{Timofte-ICCV-2013} & SRCNN\cite{Dong-ECCV-2014} & A+\cite{Timofte-ACCV-2014} & JOR\cite{JOR:EG15} & SRF\cite{SR_forest} \\
 \includegraphics[width=0.12\textwidth]{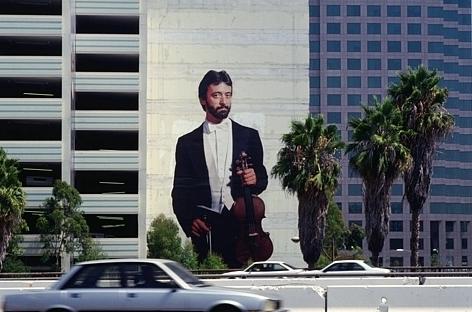} &
 \includegraphics[width=0.12\textwidth]{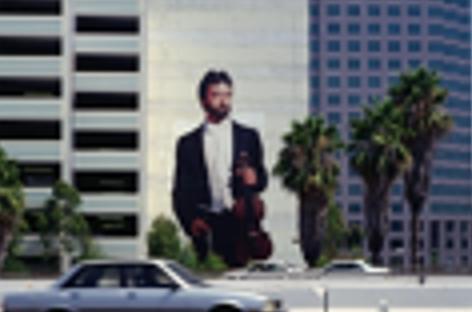} &
 \includegraphics[width=0.12\textwidth]{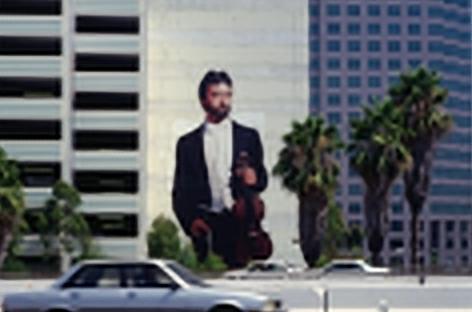} &
    \includegraphics[width=0.12\textwidth]{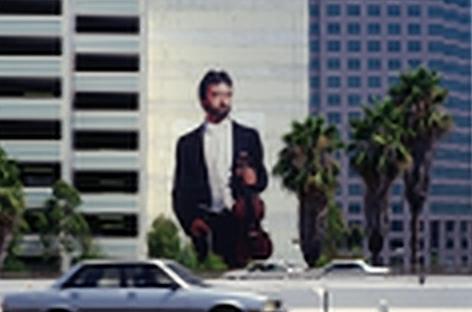} &
 \includegraphics[width=0.12\textwidth]{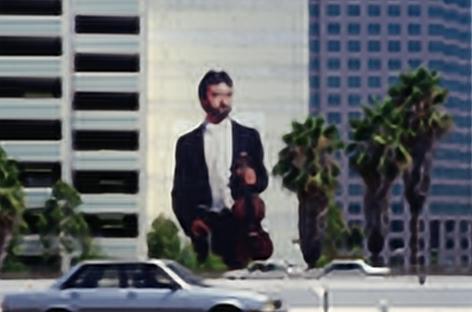} &
    \includegraphics[width=0.12\textwidth]{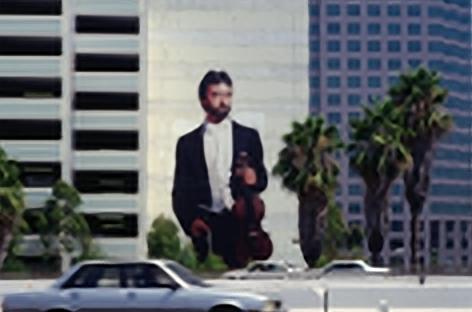} &
    \includegraphics[width=0.12\textwidth]{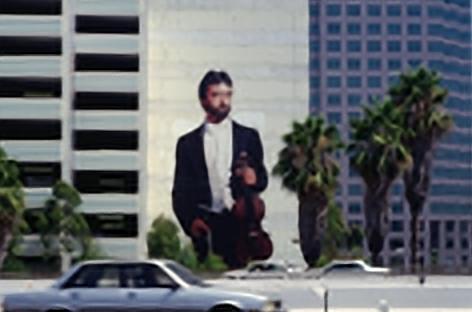} &
    \includegraphics[width=0.12\textwidth]{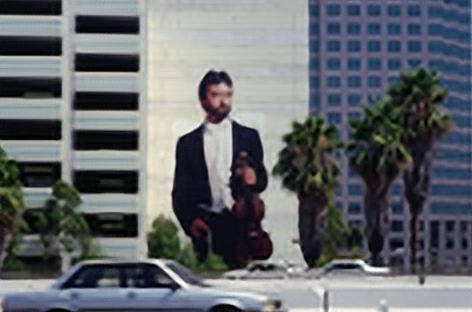} \\
    PSNR & 22.06 & 22.83 & 22.69 & 23.13 & 23.16 & 23.13 & 23.13 \\
 \includegraphics[width=0.12\textwidth]{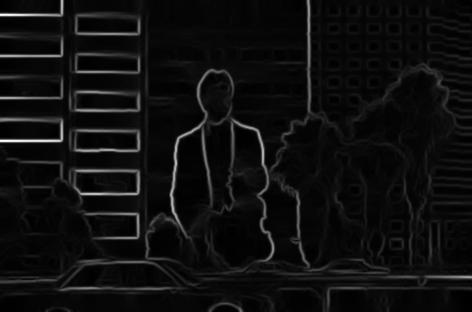} &
 \includegraphics[width=0.12\textwidth]{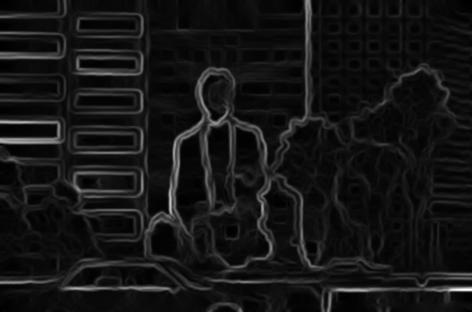} &
 \includegraphics[width=0.12\textwidth]{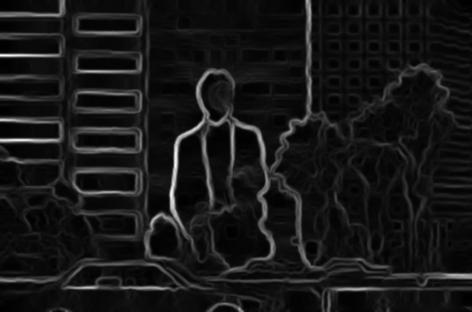} &
 \includegraphics[width=0.12\textwidth]{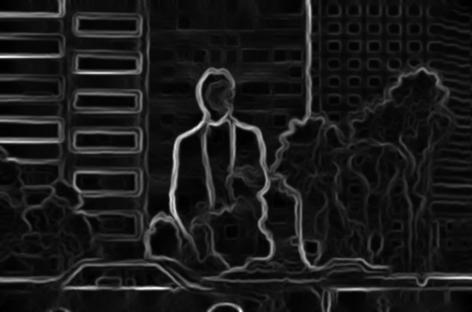} &
 \includegraphics[width=0.12\textwidth]{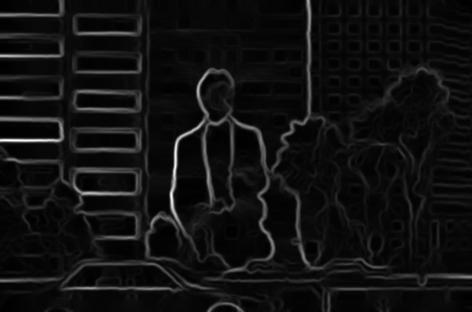} &
 \includegraphics[width=0.12\textwidth]{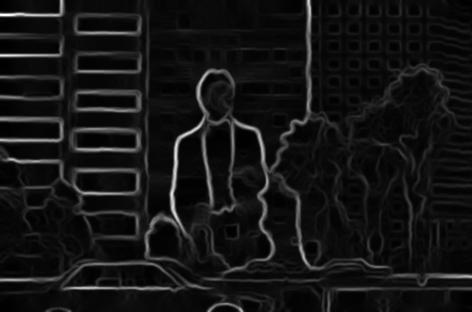} &
 \includegraphics[width=0.12\textwidth]{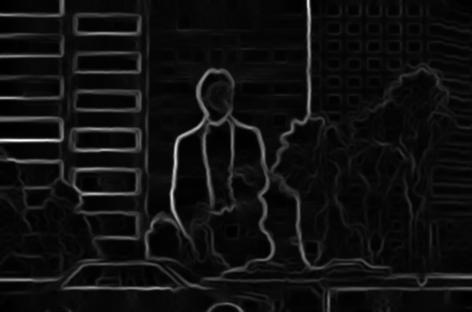} &
 \includegraphics[width=0.12\textwidth]{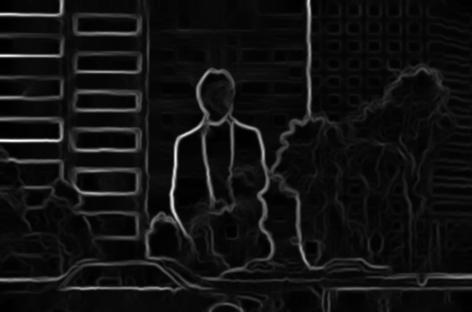} \\
    AUC & 0.718 & 0.779 & 0.739 & 0.823 & 0.807 & 0.825 & 0.828 \\
 \includegraphics[width=0.12\textwidth]{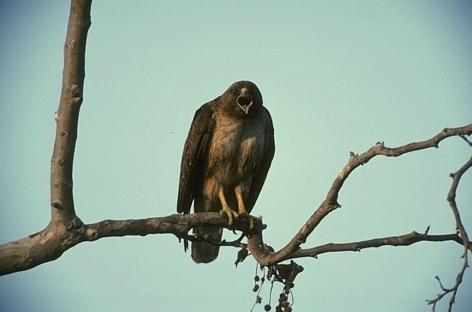} &
 \includegraphics[width=0.12\textwidth]{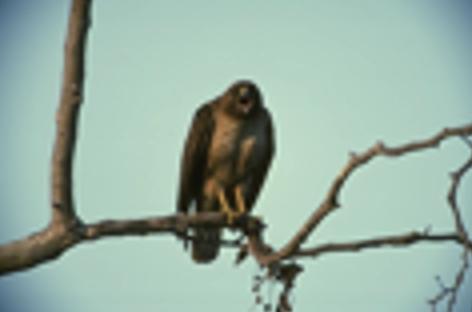} &
 \includegraphics[width=0.12\textwidth]{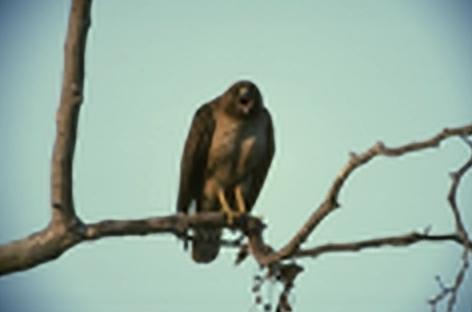} &
    \includegraphics[width=0.12\textwidth]{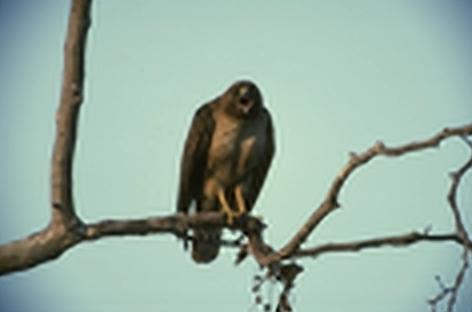} &
 \includegraphics[width=0.12\textwidth]{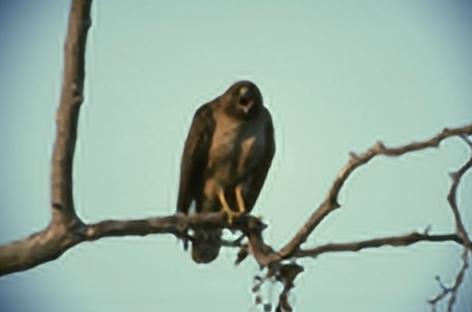} &
    \includegraphics[width=0.12\textwidth]{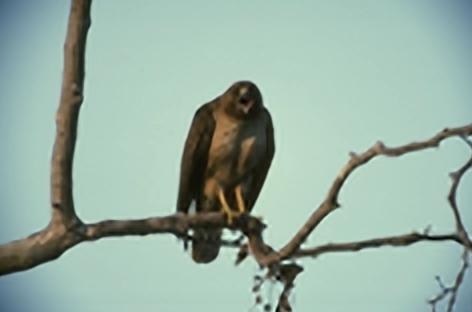} &
    \includegraphics[width=0.12\textwidth]{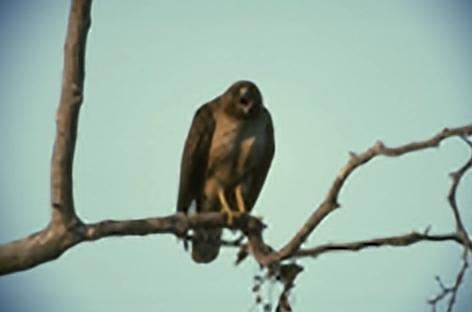} &
    \includegraphics[width=0.12\textwidth]{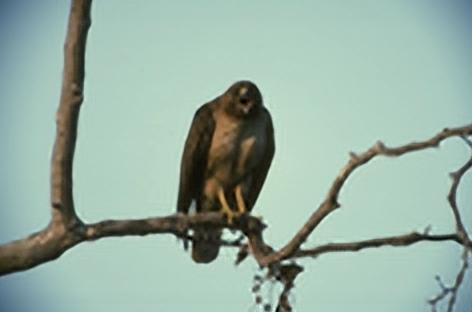} \\
    PSNR & 26.94 & 28.29 & 28.06 & 29.05 & 29.17 & 28.93 & 29.23 \\
 \includegraphics[width=0.12\textwidth]{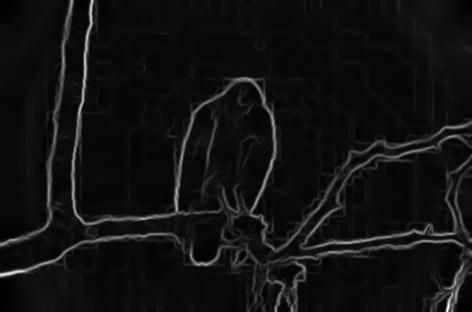} &
 \includegraphics[width=0.12\textwidth]{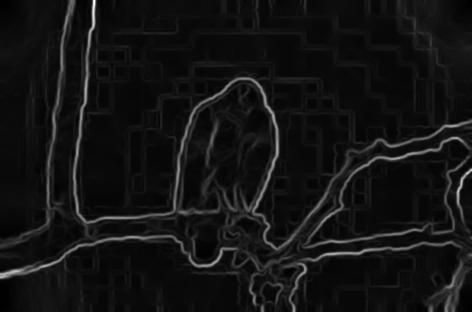} &
 \includegraphics[width=0.12\textwidth]{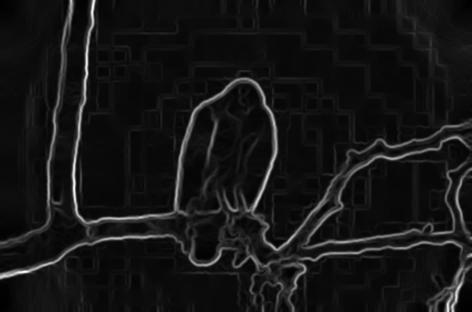} &
 \includegraphics[width=0.12\textwidth]{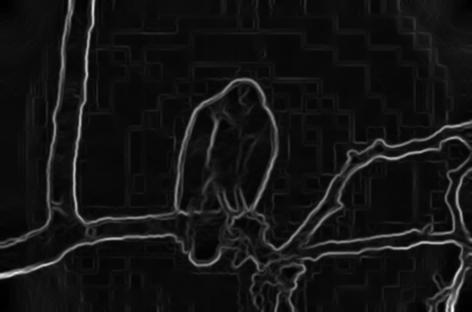} &
 \includegraphics[width=0.12\textwidth]{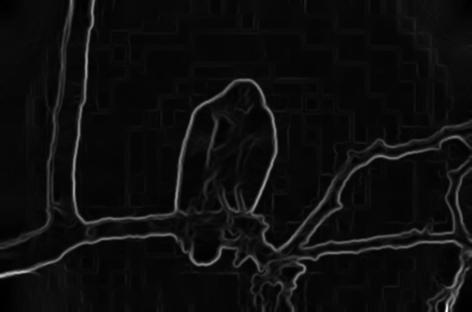} &
 \includegraphics[width=0.12\textwidth]{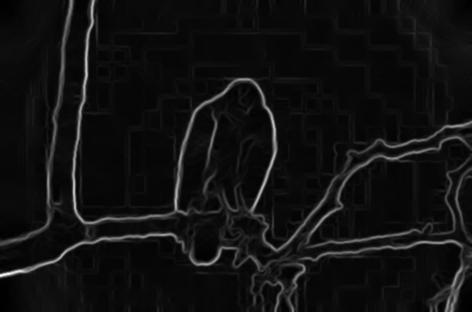} &
 \includegraphics[width=0.12\textwidth]{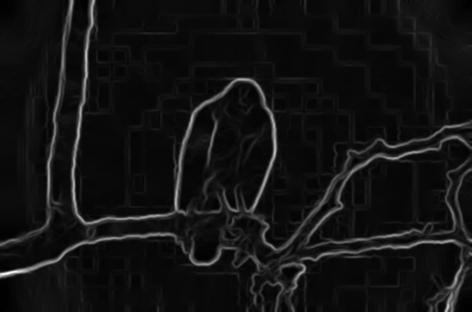} &
 \includegraphics[width=0.12\textwidth]{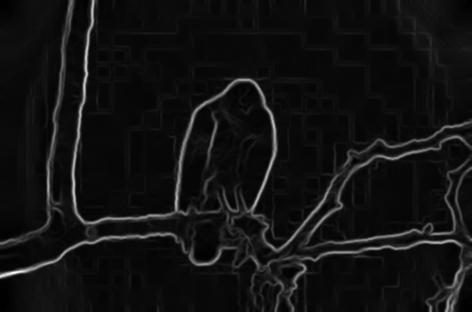} \\
    AUC & 0.861 & 0.872 & 0.870 & 0.913 & 0.900 & 0.885 & 0.891 \\
\end{tabular*}
   \caption{Super-resolved examples with their PSNR values and corresponding 
detected boundary maps by CBD~\cite{isola2014crisp} with their AUC values. 
Better seen on the screen. }
\label{fig:ed_cbd}
\end{figure*}

Table~\ref{tab:ed} lists the AUC values of BD on the eight sets of
images, along with the values of PSNR, SSIM, IFC, and NQM of
corresponding ISR methods. Fig.~\ref{fig:ed_method} shows the average
PR curves. From the table and the figure, it can be observed that ISR
methods do improve, over simple interpolation, the performance of BD
when input images are of low-resolution. For instance, JOR improves
the AUC by 0.06 when factor x4 is considered. This is because ISR
methods perform better in increasing the resolution of the LR images
to the resolution for which the BD method (CBD~\cite{isola2014crisp}
in this case) was designed. CBD uses highly localized features to
predict pixel-level boundaries, whose accuracy is affected largely by
the recovered details locally. As a result, the six learning-based
ISR methods all perform better than Bicubic Interpolation.  This
suggests that ISR should be considered as a pre-processing step for 
BD if the input images are of LR. One may argue that adapting or
re-training the BD method may increase its performance for LR images.
It is true, but we have to admit that adapting or re-training the
approach requites expertise of BD and deep understanding of the
approach used. Enhancing the resolution of LR inputs, however, is much
more straightforward for general practitioners, especially given the fact that BD is just one of
such examples as shown in following sections.

It can also be found that the four standard perceptual criteria
correlate quite well with the usefulness of ISR methods for the task
of BD. ISR methods which yield better perceptual quality (based on the
four perceptual criteria) often obtain better boundary detection
results.  However, perceptual criteria should not be considered as full proxies for the
 usefulness of ISR methods to BD. For
instance, SRCNN outscores A+ in terms of PSNR while having a lower AUC
value, when factor $\times$3 is used. This suggests that measuring the
usefulness of ISR methods for BD directly in a real system is
necessary if a high precision is requited. In general, SRCNN, A+, JOR and SRF are 
among the most useful ISR methods for the task of BD for
the dataset and approach considered. The third finding from the table
and figure is that ISR methods are more useful when the scaling factor
is larger, which means they are more needed when the input images
are of very low-resolution.

In Fig.~\ref{fig:ed_cbd}, we show visual examples, with the
super-resolution results and their corresponding BD results. From the
figure, it is evident that example-based ISR methods improve the
quality of BD results with sharper true boundaries and fewer spurious
ones. However, there is still a large room for improvement as the OB
results on the super-resolved images by the ISR methods are still
substantially worse than the result on the original (`HR') image.

\subsection{Semantic Image Segmentation}
\label{sec:il}

\begin{figure*} [tb]
\begin{tabular*}{0.4\textwidth}{ccccccccc}
\scriptsize{
(a) Original} & \scriptsize{(b) Bicubic} & \scriptsize{(c) Zeyde\cite{Zeyde-CS-2012}} & \scriptsize{(d) ANR\cite{Timofte-ICCV-2013}}
 & \scriptsize{(e) SRCNN\cite{Dong-ECCV-2014}} & \scriptsize{(f) A+\cite{Timofte-ACCV-2014}} & \scriptsize{(g) JOR\cite{JOR:EG15}} & \scriptsize{SRF\cite{SR_forest}}\\
\hspace{-2mm}
\includegraphics[width=2.15cm]{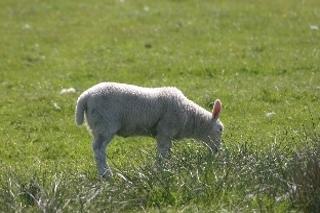} &\hspace{-5mm}
\includegraphics[width=2.15cm]{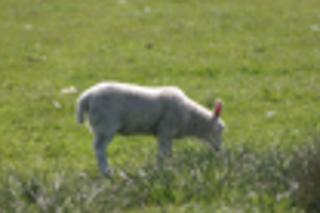} &\hspace{-5mm}
\includegraphics[width=2.15cm]{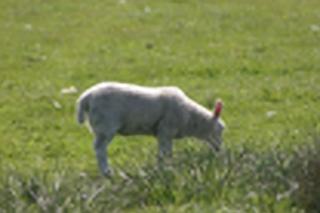} &\hspace{-5mm}
\includegraphics[width=2.15cm]{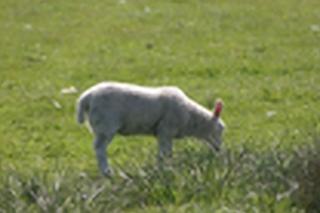} &\hspace{-5mm}
\includegraphics[width=2.15cm]{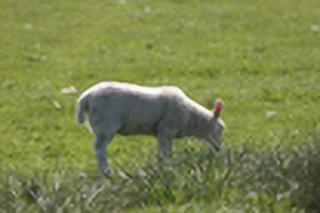} &\hspace{-5mm}
\includegraphics[width=2.15cm]{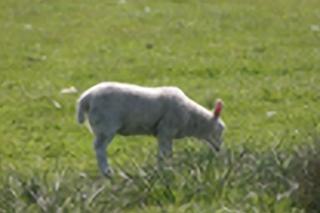} &\hspace{-5mm}
\includegraphics[width=2.15cm]{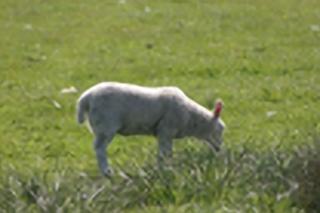} &  \hspace{-5mm}
\includegraphics[width=2.15cm]{images/1_30_s_S_lmnn_5_img.jpg}  \\
\scriptsize{---} & \scriptsize{PSNR / 26.23}  & \scriptsize{PSNR / 26.60}  & \scriptsize{PSNR / 26.64}
 & \scriptsize{PSNR /  26.65}  & \scriptsize{PSNR / {26.71}} & \scriptsize{PSNR / {26.72}} & \scriptsize{PSNR / {26.73}}\\    \hspace{-2mm}
\includegraphics[width=2.15cm]{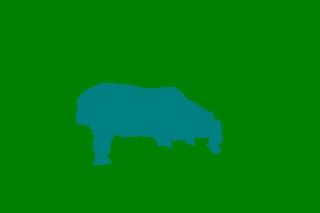} &\hspace{-5mm}
\includegraphics[width=2.15cm]{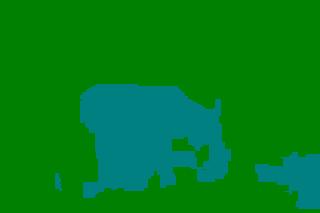} &\hspace{-5mm}
\includegraphics[width=2.15cm]{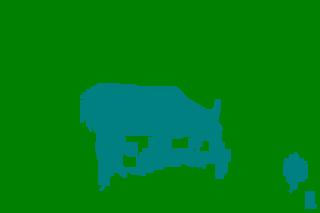} &\hspace{-5mm}
\includegraphics[width=2.15cm]{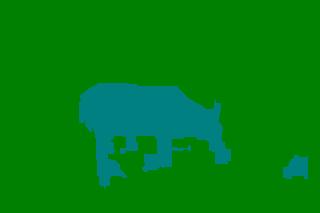} &\hspace{-5mm}
\includegraphics[width=2.15cm]{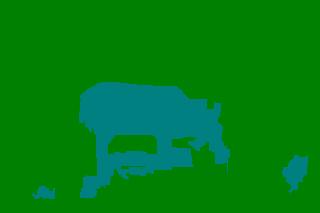} &\hspace{-5mm}
\includegraphics[width=2.15cm]{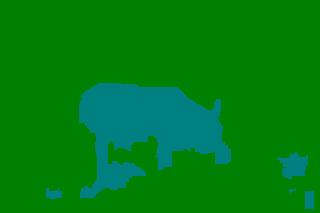} &\hspace{-5mm}
\includegraphics[width=2.15cm]{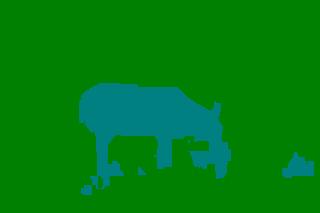} & \hspace{-5mm}
\includegraphics[width=2.15cm]{images/1_30_s_A_lmnn_5_label.jpg} & \\
\scriptsize{APP / 0.975} & \scriptsize{APP / 0.902} & \scriptsize{APP / 0.945} & \scriptsize{APP / 0.966} & \scriptsize{APP / 0.948}
 & \scriptsize{APP / 0.954} & \scriptsize{APP / 0.957} & \scriptsize{APP / 0.955}\\
\hspace{-2mm}
\includegraphics[width=2.15cm]{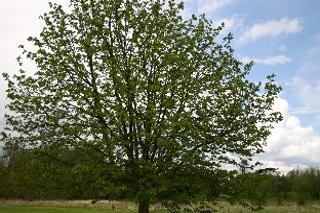} &\hspace{-5mm}
\includegraphics[width=2.15cm]{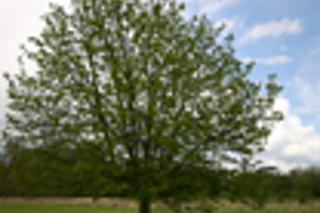} &\hspace{-5mm}
\includegraphics[width=2.15cm]{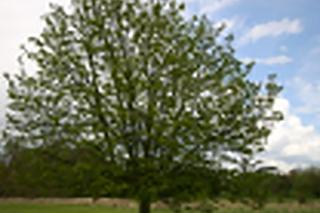} &\hspace{-5mm}
\includegraphics[width=2.15cm]{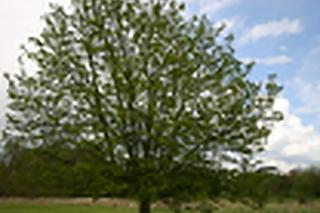} &\hspace{-5mm}
\includegraphics[width=2.15cm]{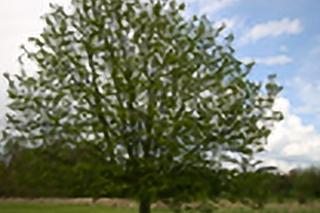} &\hspace{-5mm}
\includegraphics[width=2.15cm]{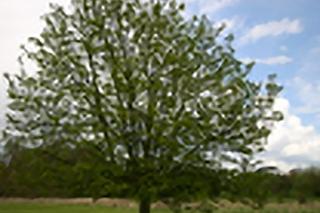} & \hspace{-5mm}
\includegraphics[width=2.15cm]{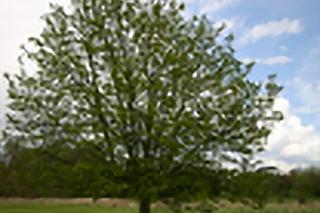} & \hspace{-5mm}
\includegraphics[width=2.15cm]{images/2_16_s_S_lmnn_5_img.jpg} &\hspace{-5mm} \\
\scriptsize{---} & \scriptsize{PSNR / 24.03}  & \scriptsize{PSNR /  24.55}  & \scriptsize{PSNR /  24.53}
& \scriptsize{PSNR /   24.83}  & \scriptsize{PSNR /  24.78} & \scriptsize{PSNR / 24.76} &\scriptsize{ PSNR / 24.78}\\    \hspace{-2mm}
\includegraphics[width=2.15cm]{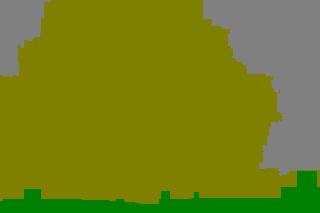} &\hspace{-5mm}
\includegraphics[width=2.15cm]{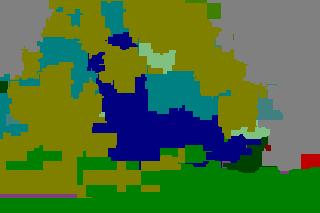} &\hspace{-5mm}
\includegraphics[width=2.15cm]{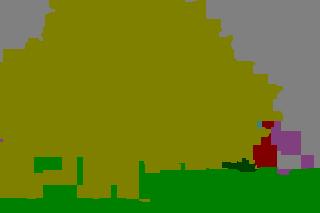} &\hspace{-5mm}
\includegraphics[width=2.15cm]{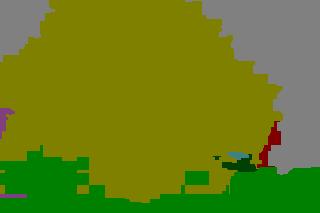} &\hspace{-5mm}
\includegraphics[width=2.15cm]{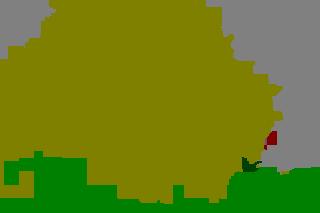} &\hspace{-5mm}
\includegraphics[width=2.15cm]{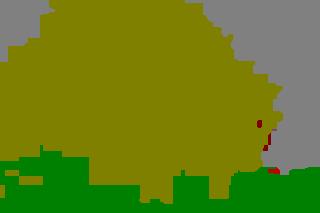} &\hspace{-5mm}
\includegraphics[width=2.15cm]{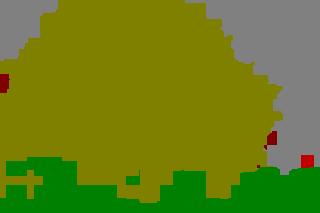} & \hspace{-5mm}
\includegraphics[width=2.15cm]{images/2_16_s_R_lmnn_5_label.jpg}  \\
\scriptsize{APP / 0.985} & \scriptsize{APP / 0.647} & \scriptsize{APP / 0.947} & \scriptsize{APP / 0.971}
 & \scriptsize{APP / 0.972} & \scriptsize{APP / 0.968} & \scriptsize{APP / 0.976}  & \scriptsize{APP / 0.978}  \\

\hspace{-2mm}
\includegraphics[width=2.15cm]{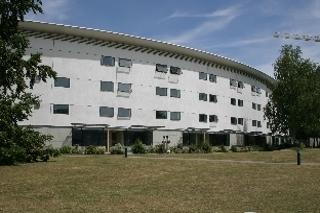} & \hspace{-5mm}
\includegraphics[width=2.15cm]{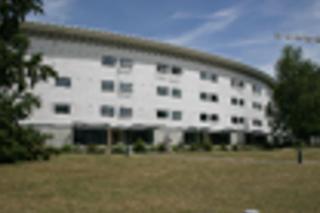} &\hspace{-5mm}
\includegraphics[width=2.15cm]{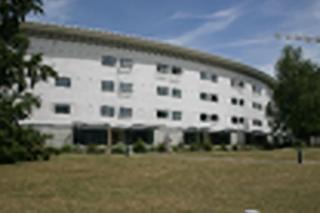} &\hspace{-5mm}
\includegraphics[width=2.15cm]{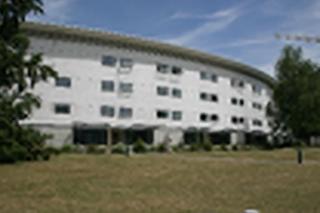} &\hspace{-5mm}
\includegraphics[width=2.15cm]{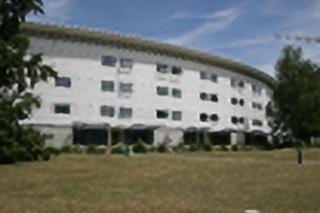} &\hspace{-5mm}
\includegraphics[width=2.15cm]{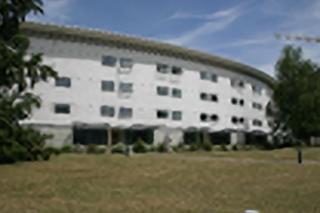} &\hspace{-5mm}
\includegraphics[width=2.15cm]{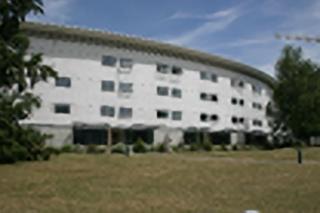} & \hspace{-5mm}
\includegraphics[width=2.15cm]{images/3_22_s_R_lmnn_5_img.jpg} \\
\scriptsize{---}  & \scriptsize{PSNR / 17.49} & \scriptsize{PSNR /   17.80} & \scriptsize{PSNR /   17.83}
  & \scriptsize{PSNR /  17.88}  & \scriptsize{PSNR / 17.85}  & \scriptsize{PSNR /  17.85}  & \scriptsize{PSNR /  17.85} \\ \hspace{-2mm}
\includegraphics[width=2.15cm]{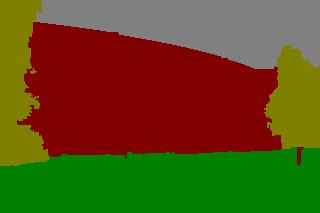} &\hspace{-5mm}
\includegraphics[width=2.15cm]{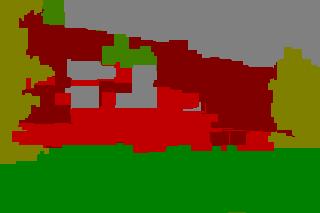} &\hspace{-5mm}
\includegraphics[width=2.15cm]{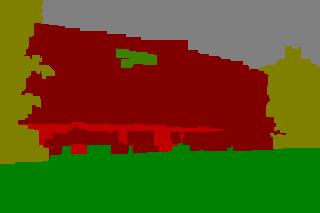}& \hspace{-5mm}
\includegraphics[width=2.15cm]{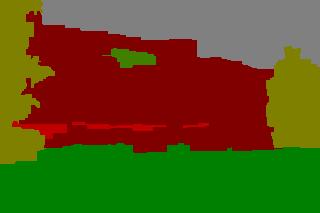} &\hspace{-5mm}
\includegraphics[width=2.15cm]{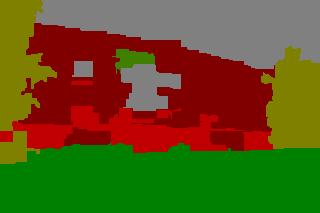} &\hspace{-5mm}
\includegraphics[width=2.15cm]{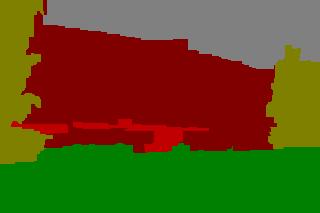} &\hspace{-5mm}
\includegraphics[width=2.15cm]{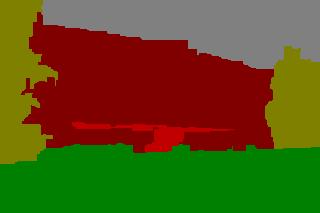} &\hspace{-5mm}
\includegraphics[width=2.15cm]{images/3_22_s_A_lmnn_5_label.jpg}  \\
\scriptsize{APP / 0.937} & \scriptsize{APP / 0.639} & \scriptsize{APP / 0.891} & \scriptsize{APP / 0.899}
& \scriptsize{APP / 0.754} & \scriptsize{APP / 0.888} & \scriptsize{APP / 0.909} & \scriptsize{APP / 0.907} \\

\end{tabular*}
\caption{Examples for semantic image segmentation: super-resolved images with their
  PSNR values and the corresponding labeling results
  with their average precision over pixels (APP) are shown. Better seen on the screen.}
\label{fig:example:il}
\end{figure*}

\begin{table*} [tb]
\centering
\caption{Average PSNR, SSIM, IFC, NQM, and labeling accuracy on MSRC-21 dataset, 
where APP indicates Average Precision over Pixels, and APC means Average Precision over Classes. 
The best performance is shown in \textbf{bold} and the second best is \underline{underlined}.}
\resizebox{0.9\textwidth}{!}
{
\begin{tabular}{|l|c|cccccc|c|c|c|}
\hline
 \multicolumn{2}{|c|}{MSRC-21} & Bicubic & Zeyde~\etal\cite{Zeyde-CS-2012} & ANR\cite{Timofte-ICCV-2013} & SRCNN\cite{Dong-ECCV-2014} & A+\cite{Timofte-ACCV-2014} & JOR\cite{JOR:EG15} & SRF\cite{SR_forest} & Original \\
\hline\hline
\textbf{$\times$3} & PSNR & 25.29 & 26.02 & 26.00 & {26.21} & \underline{26.28} & \underline{26.28} & \textbf{26.35} & --- \\
            & SSIM & 0.689 & 0.726 & 0.728 & {0.733} & \underline{0.737} & \underline{0.737} & \textbf{0.738} & --- \\
            & IFC & 2.677 & 3.214 & 3.250 & 3.131 & {3.390} & \underline{3.396} & \textbf{3.640} & --- \\
            & NQM & 19.56 & 22.48 & 22.47 & 22.64 & {23.10} & \underline{23.16} & \textbf{23.20} & --- \\
            \hline
            & APP & 0.692 & 0.762 & 0.770 & 0.777 & {0.780}  & \textbf{0.783} & \underline{0.782} &0.844 \\
            & APC & 0.592 & 0.662 & 0.674 & 0.681 & {0.684}  & \textbf{0.687} &\underline{ 0.685} & 0.743 \\
\hline

\textbf{$\times$4} & PSNR & 24.04 & 24.65 & 24.63 & 24.77 & \underline{24.88} & {24.86} & \textbf{24.90} & --- \\
            & SSIM & 0.608 & 0.641 & 0.643 & 0.646 & \underline{0.654} & {0.652} & \textbf{0.660} & --- \\
            & IFC & 1.694 & 2.043 & 2.066 & 1.992 & \underline{2.171} & {2.151} & \textbf{2.301} & --- \\
            & NQM & 14.75 & 16.56 & 16.55 & 16.73 & \underline{17.10} & \textbf{17.12} & 16.99 & --- \\
            \hline
            & APP & 0.582 & 0.665 & 0.677 & 0.673 & \textbf{0.682} & \underline{0.674} & \underline{0.674} & 0.844 \\
            & APC & 0.505 & 0.569 & 0.584 & {0.588} & \underline{0.591} & 0.586 & \textbf{0.605} & 0.743 \\
\hline
\end{tabular}
\label{tab:il}}
\end{table*}


In this section, we consider the task of semantic image segmentation, which aims to assign a semantic label to
each pixel of the image, such as \emph{tree}, \emph{road}, and
\emph{car}. It is a very popular high-level vision task with a large
number of methods proposed~\cite{texton-forests, superpixel:eccv14, fully_cnn}.  We follow
the footsteps of most previous works on semantic image segmentation and choose the
standard MSRC-21~\cite{Shotton:2006:ECCV} dataset for the
evaluation. 
MSRC-21 consists of 591 images of 21 semantic
categories. For the segmentation method, we employ the recent
approach~\cite{superpixel:eccv14} for its simplicity in order to better 
show the influence of ISR. \cite{superpixel:eccv14}
presents a fast approximate nearest neighbor algorithm for image
labeling. They build a super-pixel graph from annotated set of
training images. At test time, they transfer labels from the training
images to the test image via matching super-pixels in the graph. The
distance between super-pixels in the feature space is approximated by
edge distance in the super-pixel graph where the edge weights are
learned from the training set. This method shows comparable results to
the state-of-the-art methods. For the implementation, we use the
authors' code with the default settings.

\begin{table*} [tb]
\centering
\resizebox{\textwidth}{!}{
\begin{tabular}{|l|c|ccccccc|c|}
  \hline
  \multicolumn{2}{|c|}{SVHN} & Bicubic & Zeyde~\etal\cite{Zeyde-CS-2012} & ANR\cite{Timofte-ICCV-2013} & SRCNN\cite{Dong-ECCV-2014} & A+\cite{Timofte-ACCV-2014} & JOR\cite{JOR:EG15} & SRF\cite{SR_forest} & Original \\
  \hline
  \hline
  $\times$3 & PSNR & 33.39 & \underline{35.40} & \textbf{35.73} & 35.03 & 34.85 & 34.90 & 34.82 & --- \\
  & SSIM & 0.912 & 0.946 & \textbf{0.949} & 0.946 & 0.946 & \underline{0.948} & \underline{0.948} & --- \\
  & IFC & 2.050 & 2.331 & \textbf{2.417} & 2.291 & \underline{2.389} & 2.346 & 2.355 & --- \\
  & NQM & 10.23 & \underline{12.59} & \textbf{12.91} & 12.16 & 12.17 & 12.21 & 12.19 & --- \\   \hline
  & Accuracy & 0.766 & 0.774 & 0.777 & \textbf{0.779} & \underline{0.778} & 0.775 & \underline{0.778} & 0.793 \\
  \hline   \hline
  $\times$3[R] & PNSR & 33.39 & 36.30 & 36.53 & 35.99 & \underline{37.20} & \textbf{37.26} & 37.12 & --- \\
  & SSIM & 0.912 & 0.951 & 0.953 & 0.947 & \underline{0.963} & \textbf{0.964} & \underline{0.963} & --- \\
  & IFC & 2.050 & 2.484 & 2.550 & 2.427 & \underline{2.726} & \textbf{2.730} & 2.701 & --- \\
  & NQM & 10.23 & 13.30 & 13.53 & 13.04 & \underline{14.24} & \textbf{14.25} & 14.17 & --- \\   \hline
  & Accuracy & 0.766 & 0.775 & 0.774 & 0.773 & \underline{0.783} & \textbf{0.786} & 0.778 & 0.793 \\
  \hline   \hline
  $\times$4 & PSNR & 29.08 & 30.63 & \underline{30.72} & \textbf{30.83} & 30.45 & 30.47 & 30.11 & --- \\
  & SSIM & 0.787 & 0.842 & 0.847 & \underline{0.849} & 0.847 & \textbf{0.850} & 0.845 & --- \\
  & IFC & 1.262 & 1.352 & \underline{1.367} & \textbf{1.368} & 1.365 & 1.339 & 1.287 & --- \\
  & NQM & 6.211 & 7.864 & \underline{7.978} & \textbf{8.021} & 7.732 & 7.720 & 7.453 & --- \\   \hline
  & Accuracy & 0.712 & \underline{0.731} & \underline{0.731} & 0.730 & \textbf{0.737} & 0.722 & 0.729 & 0.795 \\
  \hline   \hline
  $\times$4[R] & PNSR & 29.08 & 31.07 & 31.06 & 31.00 & \textbf{32.00} & 31.71 & \underline{31.77} & --- \\
  & SSIM & 0.787 & 0.858 & 0.862 & 0.856 & \textbf{0.887} & \textbf{0.887} & \underline{0.886} & --- \\
  & IFC & 1.262 & 1.456 & 1.466 & 1.427 & \textbf{1.639} & \underline{1.599} & 1.589 & --- \\
  & NQM & 6.211 & 8.268 & 8.286 & 8.209 & \textbf{9.162} & 8.872 & \underline{8.962} & --- \\   \hline
  & Accuracy & 0.712 & 0.735 & 0.730 & 0.732 & \underline{0.744} & \underline{0.744} & \textbf{0.749} & 0.795 \\
  \hline
\end{tabular}
}
\caption{Results of digit recognition on the SVHN dataset. 
The $k$-NN classifier is trained and applied on HOG features of each pair of super-resolved 
training and test sets. Methods marked with [R] are retrained using the unused digits of 
the SVHN dataset. The best performance is shown in \textbf{bold} and the second best is \underline{underlined}.}
\label{tab:dr}
\end{table*}

In order to evaluate the ISR methods for semantic image segmentation, we train the
method~\cite{superpixel:eccv14} with the original training images (\eg the HR images) and
test the trained model on eight versions of the testing images,
created by down-sampling the original images and then up-solving them
by the ISR methods to the resolution of the original images. Again,
the performance is tested for scaling factor x3 and x4.
Table~\ref{tab:il} lists the results of all ISR methods, where the
average precision over pixels (APP) and the average precision over
classes (APC) are reported, along with the values of the four
perceptual criteria. As we can see from the table, all the six ISR
methods yield significantly better results than Bicubic
Interpolation. Putting it into another way, these learning-based
super-resolution systems, in addition to improving visual quality of
LR images, do facilitate semantic labeling tasks and improve the
performance substantially when the resolution of the testing images
are lower than that of the training images. The results suggest that
it is worth effort to integrate ISR methods into real image
labeling systems if the resolutions of training and testing images are
distinctive. This is highly probably the case for real semantic labeling
systems where training images on the server side are from expensive
sensors and testing images on the user side are from cheap sensors
such as cameras of an mobile phone. Another observation from the table
is that the standard perceptual evaluation criteria  correlate
quite well with the usefulness of ISR methods for semantic image segmentation. This
implies that good visual quality also facilitates computer systems for
recognition. This can ascribed to the fact that the semantics are
defined by human and computer are trained to conduct a human vision
task which is of course very relevant to the perceptual quality of
images. Also, ISR methods are more useful when the scaling factor is
larger, which means they are more needed when the input images are
of very low-resolution.  The observation is consistent with the one we had for
BD in Sec.~\ref{sec:ed}.

In Fig.~\ref{fig:example:il}, we show three image examples, with the
super-resolution results and their corresponding labeling
results. From the figure, it is evident that ISR methods improve the
quality of the labeling results. For instance, in the third example,
results of Bicubic Interpolation labeled a large area of the building
to sky, which is probably due to the detailed textures on the building
are missing in the interpolated image. The missing texture  are recovered (to some
extend) by the example-based ISR methods, leading to better labeling
results. Also, it can be found that RGB images that have small
difference in perception may lead to totally different labeling
results, e.g. the tree in the second example. This implies that there
is still room for computer recognition systems to improve
in order to be as robust as human vision.  


\subsection{Digit Recognition}
\label{sec:dr}

In this section, we test the usefulness of ISR methods for the task of
digit recognition where the training images and the test image are
both of low-resolution.  We use the Street View House Numbers
(SVHN)~\cite{37648} dataset which contains more than 100,000 images of
house numbers obtained from Google Street View. Each image presents a
single digit at its center and has the same size of 32$\times$32
pixels.  We select 26,032 and 10,000 images from the dataset as our
training and test set.  In order to evaluate the usefulness of ISR
methods for digit recognition, we here down-sample all the images by
factor x3 and factor x4, and up-sample the down-sampled images to the
resolution of the original images by the ISR methods.  As the SVHN
dataset merely presents numbers from 0 to 9, it is highly specific and
quite different from the training dataset from~\cite{Yang-TIP-2010}
that is used to train the ISR methods. Therefore, we re-trained all ISR methods
with the unused images from the SVHN dataset, to study the generality of ISR methods. After adding the
re-trained methods, we now have twelve datasets of super-resolved
results, one dataset from Bicubic Interpolation and one dataset of
the original images. As to the classifier, we use the $k$-NN with $k=5$ for each of the eight
image sets with HOG feature~\cite{Dalal_HoG} as input. Other values of $k$ yield a similar trend.

\begin{figure*}
\centering
\resizebox{0.95\textwidth}{!}{
\setlength{\tabcolsep}{3pt}
\begin{tabular}{cccccccc}
    Original & Bicubic & Zeyde\cite{Zeyde-CS-2012} & ANR\cite{Timofte-ICCV-2013} & SRCNN\cite{Dong-ECCV-2014} & A+\cite{Timofte-ACCV-2014} & JOR\cite{JOR:EG15} & SRF\cite{SR_forest} \\
     \includegraphics[width = 0.15\textwidth]{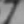} &
        \includegraphics[width = 0.15\textwidth]{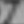} &
    \includegraphics[width = 0.15\textwidth]{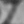} &
    \includegraphics[width = 0.15\textwidth]{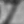} &
    \includegraphics[width = 0.15\textwidth]{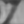} &
    \includegraphics[width = 0.15\textwidth]{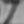} &
    \includegraphics[width = 0.15\textwidth]{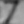} &
    \includegraphics[width = 0.15\textwidth]{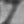} \\
    & 27.83& 29.02 & 29.59 & 29.37 & 31.37 & 31.51 & 29.93 \\
       \huge{\text{re-trained}} &
        \includegraphics[width = 0.15\textwidth]{././fig/dr/test_05542_2-Bicubic_.png} &
    \includegraphics[width = 0.15\textwidth]{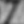} &
    \includegraphics[width = 0.15\textwidth]{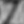} &
    \includegraphics[width = 0.15\textwidth]{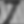} &
    \includegraphics[width = 0.15\textwidth]{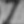} &
    \includegraphics[width = 0.15\textwidth]{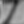} &
    \includegraphics[width = 0.15\textwidth]{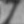} \\
    & 27.83 & 31.46[R] & 30.50[R] & 30.46[R] & 32.81[R] & 32.84[R] & 33.75[R] \\

        \includegraphics[width = 0.15\textwidth]{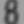} &
    \includegraphics[width = 0.15\textwidth]{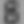} &
    \includegraphics[width = 0.15\textwidth]{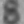} &
    \includegraphics[width = 0.15\textwidth]{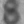} &
    \includegraphics[width = 0.15\textwidth]{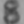} &
    \includegraphics[width = 0.15\textwidth]{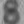} &
    \includegraphics[width = 0.15\textwidth]{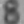} &
    \includegraphics[width = 0.15\textwidth]{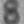} \\
    & 25.44 & 26.59 & 26.27 & 26.86 & 26.77 & 27.19 & 26.63 \\
    \huge{\text{re-trained}} &
        \includegraphics[width = 0.15\textwidth]{././fig/dr/test_05369_2-Bicubic_.png} &
    \includegraphics[width = 0.15\textwidth]{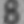} &
    \includegraphics[width = 0.15\textwidth]{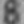} &
    \includegraphics[width = 0.15\textwidth]{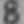} &
    \includegraphics[width = 0.15\textwidth]{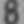} &
    \includegraphics[width = 0.15\textwidth]{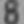} &
    \includegraphics[width = 0.15\textwidth]{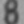} \\
     & 25.44 & 29.24[R] & 27.87[R] & 27.71[R] & 33.00[R] & 32.82[R] & 35.58[R] \\
\end{tabular}
}
\caption{Super-resolved results and corresponding PSNR values of two digits. 
Methods marked with [R] are retrained using the unused digits of the SVHN dataset. Better seen on the screen.}
\label{fig:dr}
\end{figure*}

The classification performance is listed in Table~\ref{tab:dr}. The table demonstrates that ISR methods do improve the
performance of digit recognition over simple interpolation, and that
the four perceptual criteria correlate quite well with the usefulness
of ISR methods for digit recognition. The reason of the improvement is
that HOG feature was designed for images of normal resolution, so by
applying ISR methods to the LR input images, HOG can be extracted from
images of suitable resolution.  However, we find that with the
standard, general training dataset~\cite{Yang-TIP-2010}, Zeyde et
al. and ANR perform better than SRCNN, A+ , JOR and SRF, which is different
from the results of the previous two tasks with general images. This observation
suggests that Zeyde et al. and ANR are more generally applied than the
other four state-of-the-art ISR methods. One possible reason is that models 
of higher complexity are more likely to overfit to the training data. 
The problem can be solved by
re-training the model with data of similar distribution as the test data. We re-trained
all the six method with unlabeled digits in SVHN, and as expected the
performance is improved significantly, according to the four perceptual criteria or 
recognition accuracy. See Table~\ref{tab:dr} for the improvement. After re-training, 
the four methods SRCNN, A+, JOR, and SRF yield the best results.  
 In Fig.~\ref{fig:dr}, we show
two digits, along with their super-resolved results by factor x3
and the PSNR values. From the figure, it is clear to see the artifacts
generated by the ISR methods trained with general training data. The
introduced artifacts lead to noisy HOG features, which in turn confuse
the classifier. All the evidence leads to conclusions similar to that
drawn for boundary detection and semantic image segmentation: (1) ISR methods are
generally helpful for recognizing digits of low-resolution; and (2)
perceptual evaluation criteria reflect the usefulness of ISR
to digit recognition quite well.  In addition, we find that the performance of
ISR methods will improve significantly if they are re-trained with domain specific
data.


\subsection{Scene Recognition}
\label{sec:scene}

\begin{table*} [tb]
 \centering
\resizebox{0.92\textwidth}{!}{
\begin{tabular}{|l|c|ccccccc|c|c|}
  \hline
 \multicolumn{2}{|c|}{Scene-15} & Bicubic & Zeyde~\cite{Zeyde-CS-2012} & ANR\cite{Timofte-ICCV-2013} & SRCNN\cite{Dong-ECCV-2014} & A+\cite{Timofte-ACCV-2014} & JOR\cite{JOR:EG15} & SRF~\cite{SR_forest}& Original \\
  \hline
 $\times$3 & PSNR & 25.12 & 25.85 &  25.87 &  26.10 &  \textbf{26.19} &  \underline{26.18} &  26.13    & --- \\
  & SSIM & 0.73 & 0.78 &  0.77 &  0.78 &  \underline{0.79} &  \underline{0.79} &  \textbf{0.80}    & --- \\
  & IFC & 2.82 & 3.34 &  3.43 &  3.20 &  \underline{3.58} &  \textbf{3.60} &  \underline{3.58}    &--- \\
  & NQM & 19.75 & 22.69 &  22.73 &  22.81 &  \underline{23.39} &  \textbf{23.43} &  23.30   & --- \\ \hline
 & Accuracy  &  0.770  &  0.777  &  0.777 & 0.780  &  \textbf{0.782}  &  \textbf{0.782} & 0.778 & 0.809  \\
  \hline \hline
  $\times$4 & PSNR & 24.32 & 24.99 &  24.95 & 25.06 &  \textbf{25.24} &  \underline{25.22} & 25.19 & --- \\
  & SSIM & 0.674 & 0.701 &  0.702 &  0.704 &  \underline{0.720} &  {0.719} & \textbf{0.722} &--- \\
  & IFC & 1.597 & 1.923 &  1.911 &  1.806 & \textbf{2.021} &  {2.010} & \underline{2.014} & --- \\
  & NQM & 14.43 & 16.12 &  16.05 &  16.07 & \textbf{16.62} &  \underline{16.61} & 16.57&  --- \\ \hline
  & Accuracy &   0.735  &  0.752  &  0.753 & 0.748 & \textbf{0.754} &  \underline{0.753}  & \underline{0.753} & 0.809   \\
  \hline
\end{tabular}
}  
\caption{Average PSNR, SSIM, IFC, NQM values and the accuracy of scene recognition on Scene-15 dataset. }
\label{tab:scene}
\vspace{-2mm}
\end{table*}

In this section, we evaluated six ISR methods on the task of scene recognition. 
We tested the methods on the Scene-15 dataset~\cite{lazebnik:cvpr06}, which has been widely 
used for image classification and clustering~\cite{lazebnik:cvpr06, dai:ensemble:eccv12, dai:EnPro:iccv13}. 
Scene-15 contains 15
scene categories in both indoor and outdoor environments. Each category has 200
to 400 images, and they are of size $300 \times 250$ pixels on average.
We use the same experimental designs as for the previous tasks:  down-sampling all the images by 
factor x3 and factor x4, and up-sampling the down-sampled images to the
resolution of the original images by the six ISR methods, thus resulting in six super-resolved datasets 
for each scaling factor, one for bicubic interpolation, and one for the original (HR) images.  
As to the features, we use the  Convolutional Neural Network (CNN) features~\cite{deep:bmvc14}, 
obtained from an off-the-shelf CNN model pretrained on the ImageNet. The feature is  chosen as CNN 
feature has achieved state-of-the-art performance for image
classification~\cite{deep:bmvc14}. It is worth noticing that the training and testing data are processed 
the same way, \ie down-sampled by bicubic interpolation and up-sampled by the same ISR method (one of the six).  
The convolutional results at layer 16 were
stacked as the CNN feature vector, with dimensionality of
4096.  As to the classification, we use $15$ images per class as the training samples, and the rest left for testing. 

The classification accuracies over 10 random training-testing splits are averaged and reported in Table~\ref{tab:scene}, along with the results according to the four perceptual criteria.  
The table shows that learning-based ISR methods are helpful for scene 
recognition with the deep neural network when the input images are of low-resolution. 
The four perceptual criteria also correlate 
generally well with usefulness of ISR methods for this task, 
which is in line with the conclusions drawn for previous vision tasks.   
Images at multiple scales have recently been employed for training deep 
neural networks~\cite{cnn:multi-scale, fully_cnn},  and 
they show improvement over a single scale.  It is interesting to 
see how ISR methods help to generate multiple scales of the input images to train 
better neural networks. We leave this as our future work.

\section{Discussion and Conclusion}
\label{sec:conclusion}
We have evaluated the usefulness of image super-resolution (ISR) for a
variety of different vision tasks.  Six ISR methods have been
employed and evaluated on four popular vision tasks. Three general
conclusions can be drawn from experiments on the four tasks: 1) ISR
methods are helpful in general for other vision tasks when the
resolution of input images are low; 2) standard perceptual criteria,
namely PSNR, SSIM, IFC, NQM, correlate quite well with the usefulness
of ISR methods for the vision tasks, but they are not
accurate enough to be used as full proxies; and 3) even with the state-of-the-art ISR methods, the
performance with the super-resolved images are still significantly
inferior to that with the original, high-resolution images.

Although it is generally believed that ISR methods is helpful for other
vision tasks, this work has formalized the common conception and
conducted quantitative evaluation.  We hope this work will be an
inspiration for the community to integrate ISR methods into other
vision systems when the input images are of low-resolution or when multiple resolutions are needed, and to
evaluate ISR methods in real vision tasks, in addition to merely
inspecting the visual quality. 
The work may inspire the community to design super-resolution algorithm
for specific vision task rather than merely levering the perceptual criteria.

We acknowledge that for some tasks, the approaches and the datasets do not represent the state of the arts.
However, they are standard ones and we believe they are sufficient to support the conclusions.   
Method evaluation on more vision tasks with more challenging datasets, 
testing multiple approaches for the same task, and testing
different parameter settings for the same approach constitute our future
work.  The code and data of this work are available at the project page.

\textbf{Acknoledgements} The work is supported by the ERC Advanced Grant Varcity (\#273940).

\small{
\bibliographystyle{ieee}
\bibliography{egbib}
}

\end{document}